%% file: sample-sigconf.tex
\documentclass[sigconf]{acmart}

\usepackage{tablefootnote}
\usepackage{times}
\usepackage{soul}
\usepackage{url}
\usepackage[utf8]{inputenc}
\usepackage{graphicx}
\usepackage{amsmath}
\usepackage{booktabs}
\usepackage{algorithm}
\usepackage{isomath}
\usepackage{amssymb}
\usepackage{footnote}   
\usepackage{graphicx}
\usepackage{subfig}
\usepackage{multirow}
\usepackage{array}
\usepackage{bbm}
\usepackage{footmisc}
\usepackage{booktabs} % For formal tables
\usepackage{graphicx}
\usepackage{epsfig}
\usepackage{booktabs} % For formal tables
\usepackage{threeparttable}
\usepackage{colortbl} % use table color
\usepackage{color}
\usepackage{textcomp}
\usepackage{soul} % to highlight text
\usepackage{amsmath}% to input mathematics symbol
\usepackage{comment}
\usepackage{multirow}
\usepackage{algorithm} %format of the algorithm
\usepackage{algpseudocode} %format of the algorithm

\setcopyright{rightsretained}
\copyrightyear{2019} 
\acmYear{2019} 
\setcopyright{acmcopyright}
\acmConference[DLP-KDD'19]{1st International Workshop on Deep Learning Practice for High-Dimensional Sparse Data}{August 5, 2019}{Anchorage, AK, USA}  
\acmPrice{15.00}
\acmDOI{10.1145/3326937.3341256}
\acmISBN{978-1-4503-6783-7/19/08}

\begin{document}
\title[AMAD: Adversarial Multiscale Anomaly Detection]{AMAD: Adversarial Multiscale Anomaly Detection on High-Dimensional and Time-Evolving Categorical Data}
% \subtitle{Extended Abstract}
% \subtitlenote{The full version of the author's guide is available as
%   \texttt{acmart.pdf} document}
%\author{%
%Zheng Gao\affmark[1], Lin Guo\affmark[2], Chi Ma\affmark[2], Xiao Ma\affmark[2], Kai Sun\affmark[2], Hang Xiang\affmark[2], Xiaoqiang Zhu\affmark[2], Li Huang\affmark[2], Hongsong Li\affmark[2]\\
%\affaddr{\affmark[1]Indiana University Bloomington}\\
%\affaddr{\affmark[2]Alibaba Group}\\
%\emailaddr{gao27@indiana.edu,...}\\%
%}

\author{Zheng Gao}
\affiliation{
   \institution{Indiana University Bloomington}}
\email{gao27@indiana.edu}

\author{Lin Guo}
\affiliation{
   \institution{Alibaba Group}}
\email{lin.gl@alibaba-inc.com}

\author{Chi Ma}
\affiliation{
   \institution{Alibaba Group}}
\email{beiji.mc@alibaba-inc.com}

\author{Xiao Ma}
\affiliation{
   \institution{Alibaba Group}}
\email{maxiao.mx@alibaba-inc.com}

\author{Kai Sun}
\affiliation{
   \institution{Alibaba Group}}
\email{luchen.sk@alibaba-inc.com}

\author{Hang Xiang}
\affiliation{
   \institution{Alibaba Group}}
\email{xingzhi.xh@alibaba-inc.com}

\author{Xiaoqiang Zhu}
\affiliation{
   \institution{Alibaba Group}}
\email{xiaoqiang.zxq@alibaba-inc.com}

\author{Hongsong Li}
\affiliation{
   \institution{Alibaba Group}}
\email{hongsong.lhs@alibaba-inc.com}

\author{Xiaozhong Liu}
\affiliation{
   \institution{Indiana University Bloomington}}
\email{liu237@indiana.edu}

% \author{Zheng Gao, Lin Guo,Chi Ma, Xiao Ma, Kai Sun, Hang Xiang, Xiaoqiang Zhu, Li Huang, Hongsong Li}
% \affiliation{%
%   \institution{Indiana University Bloomington,Alibaba Group}}
% \email{}

\renewcommand{\shortauthors}{Zheng Gao et al.}

\begin{abstract}
Anomaly detection is facing with emerging challenges in many important industry domains, such as cyber security and online recommendation and advertising.
The recent trend in these areas calls for anomaly detection on time-evolving data with high-dimensional categorical features without labeled samples. Also, there is an increasing demand for identifying and monitoring irregular patterns at multiple resolutions. In this work, we propose a unified end-to-end approach to solve these challenges by combining the advantages of Adversarial Autoencoder and Recurrent Neural Network. The model learns data representations cross different scales with attention mechanisms, on which an enhanced two-resolution anomaly detector is developed for both instances and data blocks. Extensive experiments are performed over three types of datasets to demonstrate the efficacy of our method and its superiority over the state-of-art approaches.
\end{abstract}



% We no longer use \terms command


\maketitle

\input{intro.tex}
\input{review.tex}
\input{method.tex}
\input{experiment.tex}
\input{conclusion.tex}
\bibliographystyle{ACM-Reference-Format}
\bibliography{acmart} 
\input{appendix.tex}

\end{document}

%% file: intro.tex
\section{Introduction}
\begin{figure} 
	% \advance\leftskip-1cm 
	\centering
	\includegraphics[width=1\columnwidth]{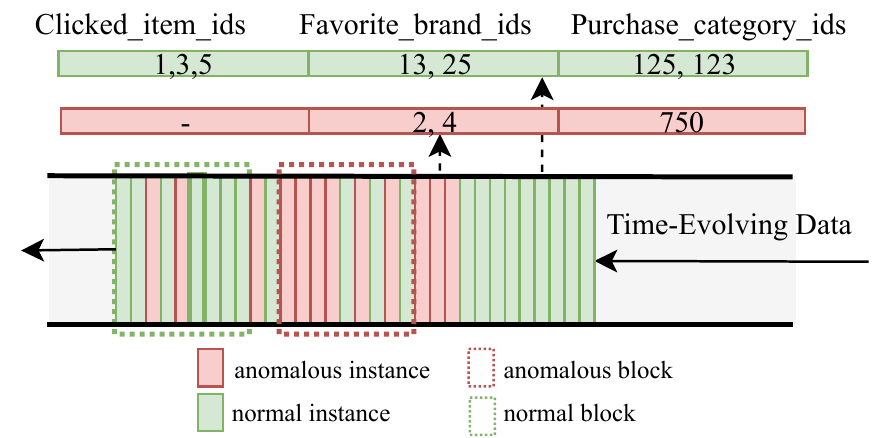}
%	\vspace{-1em}
	\caption{An example to show the data format in an online recommendation system. Each data instance contains multiple attributes composed of a group of categorical features. Either system failure or fraud attack can generate anomalous data with irregular values.}
	\label{fig:example}
%	\vspace{-1em} 
\end{figure}

Anomaly detection aims at identifying outliers or irregular patterns which are inconsistent with the majority of data. It can provide a wide range of applications, 
from capturing rare events or unusual observations to protecting a complex system against failures or attacks. Recent trend in many important industrial domains, such as online recommendation and advertising (as illustrated in Figure \ref{fig:example}), online financial service and cyber security, has set four unprecedented challenges for anomaly detection. \textbf{\emph{First}}, as data is changed with time, there is no gold standard for anomalous data across all time periods. \textbf{\emph{Second}}, labeled anomalous samples are rarely available. \textbf{\emph{Third}}, data format can be very complex, for example, a compound of attributes consisting of categorical ids with extremely high dimension. The sparse, sophisticated and noisy couplings among massive features make it very difficult to recognize the underlying patterns through handcrafted rules or feature engineering. \textbf{\emph{Fourth}}, a systematic monitoring may require detecting anomalous events at different resolutions for different needs. The data patterns can vary with scales. Although straightforward, aggregating small-scale detection results for larger-scale detection is not guaranteed to be effective. For example, detecting collective patterns such as phase distortion.

Deep learning has drawn immense attention in the field of anomaly detection. Deep neural network has the potential to automatically learn complex feature representations, thus making it possible to train an anomaly detector in an end-to-end fashion with less non-trivial expertise feature engineering.
Inspired by Generative Adversarial Network (GAN) \cite{goodfellow2014generative} and Adversarial Autoencoder \cite{makhzani2015adversarial}, some researches have been reported to adversarially train a pair of neural networks (generator and discriminator) through unsupervised or semi-supervised learning to construct an anomaly detector \cite{zenati2018adversarially,akcay2018ganomaly,sabokrou2018adversarially}. The networks' losses, especially adversarial loss and reconstruction loss, are used to identify anomalies. By design, normal samples should follow the distribution close to the majority of the training data, and thus obtain lower losses than anomalous samples.
Assuming that the training data is normal, this type of approaches do not require labeled anomalies for training. 
They are able to solve the second challenge (lack of label) listed above but do not cover the other three. 

In this paper, following the emerging idea of adversarially learned anomaly detection, we present an adversarial multiscale anomaly detector (AMAD) to tackle the aforementioned challenges in an end-to-end manner. We train a pair of deep encoder-decoder generator and discriminator to fit the normal patterns of the unlabeled training data (\textit{Challenge 2}), and using a compound loss as anomaly score for inference. We combine sequential and hierarchical representation learning to detect anomalies at two different scales (\textit{Challenge 4}) for time-evolving high-dimensional categorical data (\textit{Challenge 1,3}).
The main contributions of this work are highlighted as follows \footnote{Our codes and datasets will be made available at publication time.}:
\begin{itemize}
\item To the best of our knowledge, AMAD is the first unified end-to-end approach to tackle the aforementioned important challenges. Especially, our work is the first attempt to extend adversarial anomaly detector to the scenario of complex high-dimensional categorical data.
\item We introduce a multiscale data representation learning mechanism. Patterns are extracted and inspected cross a range of scales, from single features of an individual instance up to data blocks. This produces an enhanced two-resolution anomaly detector for both individual instances and data blocks.
\item We report extensive experiments on three types of datasets, validating that our model outperforms the state-of-arts notably. Moreover, we conduct ablation studies to prove the efficacy of the key components in our model. 
\end{itemize}

%% file: review.tex
\section{Related Works}

Deep learning has been widely applied in all research topics such as ranking \cite{gao2018end}, graph mining \cite{gao2018edge2vec}  and text generation \cite{wang2019neural}, etc. As a fundamental one, anomaly detection has been extensively studied via unsupervised or semi-supervised deep approaches.
iForest \cite{liu2012isolation}, one of the most famous approaches, utilizes a tree-based structure to split data randomly and ranks data points as anomalous based on how easy they get isolated. Affiliated with Support Vector Machine (SVM) family, one-class SVM classifiers \cite{Scholkopf:2001:ESH,chang2011libsvm,tax2004support} use designed kernels to project data to a latent space and search for a best hyperplane to set anomalies apart. 
Derived from these works, kernel-based one-class classification is further combined with deep neural network \cite{ruff2018deep,chalapathy2018anomaly} to automatically extract useful features from massive complex data. 

Deep learning attracts increasing attentions for the past decade. As a basic type of deep learning framework, autoencoder has already widely applied on anomaly detection \cite{sakurada2014anomaly,andrews2016detecting,chalapathy2017robust,zhou2017anomaly}. It learns to compress the input data with multiple hidden layers and reconstruct the input data through an encoding-decoding mechanism. Trained solely on normal data, autoencoder fails to reconstruct anomalous sample and produces large reconstruction error that can be used to identify anomaly. Furthermore, an autoencoder ensemble with adaptive sampling is proposed to improve the robustness on noisy data \cite{chen2017outlier}.

Recently, Generative Adversarial Networks rise up as a popular track in deep learning  \cite{schlegl2017unsupervised,sabokrou2018adversarially,kliger2018novelty}. Typically, a GAN-based model consists of two parts, i.e., generator and discriminator. The generator learns a representation to resemble the original input data, while the discriminator is trained to distinguish between the resembled and original inputs. The adversarial training enhances the model's ability of learning the distribution of input normal data, and is proven to be very effective for identifying anomalous or novel data.

Combining GAN and autoencoder, Adversarial Autoencoder \cite{makhzani2015adversarial} offers an alternative way for unsupervised or semi-supervised anomaly detection. Unlike GAN approaches which learns a distribution to generate discrete samples, Adversarial Autoencoder uses autoencoder as the generator to learn to resemble data.
By mapping the input to latent space and remapping back to input data space (reconstruction), it enables not only better reconstruction but also control over latent space \cite{creswell2018generative,mirza2014conditional}. Taking this track of thoughts, BiGAN \cite{donahue2016adversarial} and ALI \cite{dumoulin2016adversarially} both apply variational autoencoder as the generators in their models to optimize the distribution of normal data.
Two following works \cite{akcay2018ganomaly,zenati2018adversarially} combine both GAN and Adversarial Autoencoder components to jointly train an anomaly detector and use the reconstruction errors as the criteria to judge whether testing data is anomalous or not.

%% file: method.tex
\section{Method}

\subsection{Overview}

\begin{figure} 
	% \advance\leftskip-1cm 
	\centering
	\includegraphics[width=1.1\columnwidth]{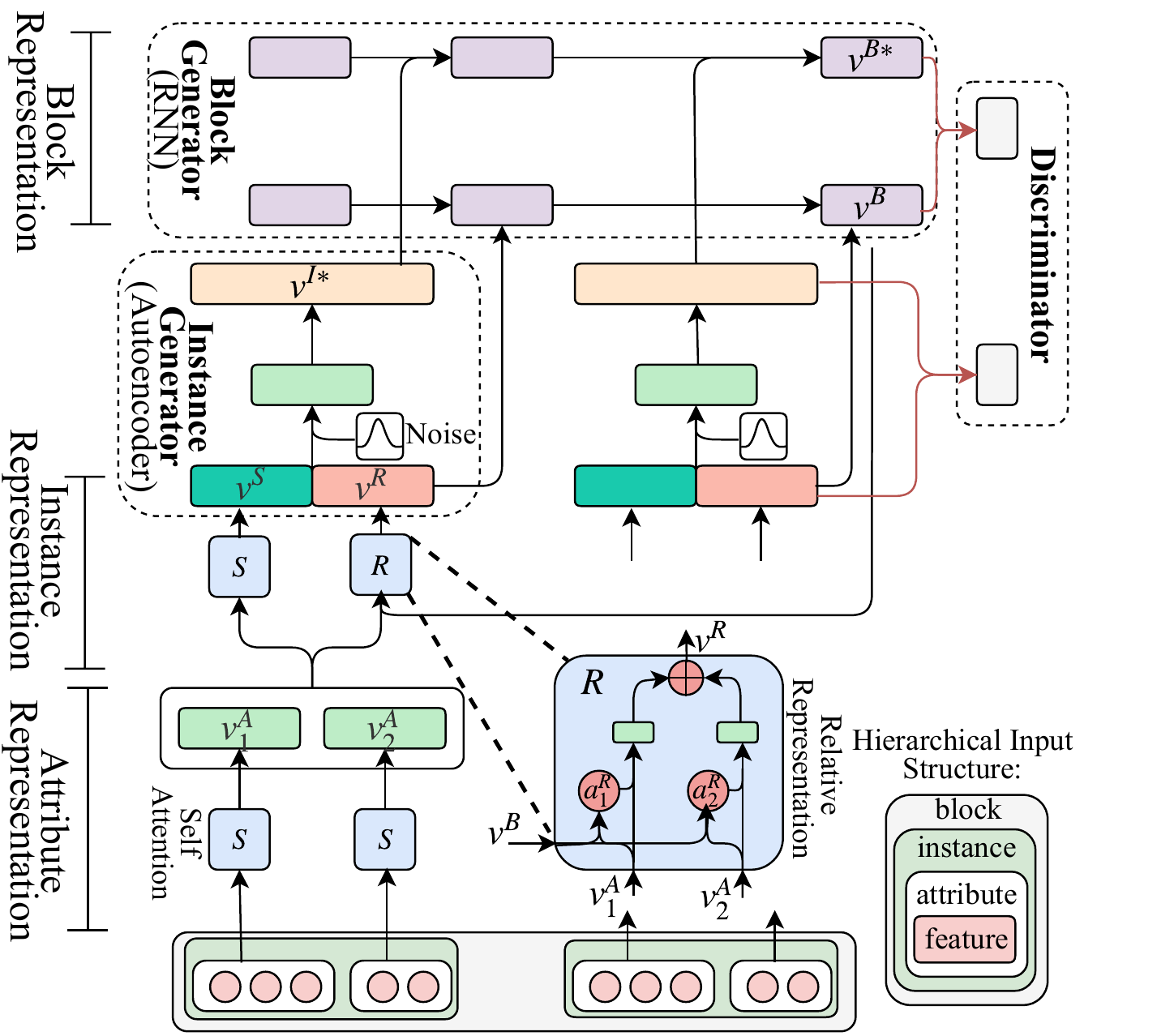}
%	\vspace{-1em}
	\caption{The overall architecture of AMAD}
	
	\label{fig:pipeline}
%	\vspace{-1em} 
\end{figure}

The framework of our model is sketched in Figure \ref{fig:pipeline}.
Our approach adversarially trains an anomaly detection model on unlabeled data, with the assumption that the training data is normal (at least mostly normal). The input data for the model is of a hierarchical four-level structure (also illustrated in Figure \ref{fig:pipeline}). We use the model to detect anomalies at the top two levels (instance and block). 
Since the model is trained to fit the distribution of normal data, the anomalies should have higher loss than normal data. Therefore, we use the loss to infer anomalies \cite{schlegl2017unsupervised,sabokrou2018adversarially,kliger2018novelty,akcay2018ganomaly,zenati2018adversarially}. 

In the following sections, we first introduce the multiscale representation learning across different levels. Second, we describe the adversarial learning architecture. In the end, we explain how we train the model and use the model for inference.

\subsection{Multiscale Representation Learning}
Our model hierarchically learns representations for the structured input data, from feature, attribute, instance, up to instance block level. We implement attention mechanism \cite{lin2017structured,vaswani2017attention,zhou2018deep} to summarize informations with distributed weights of importance to form the next-level representation, so that most important informations are extracted to the high level.

\subsubsection{Feature and Attribute Representation}
For the input layer, sparse embedding is implemented to embed each categorical feature to a fixed-size dense vector $v^{F}$, which is automatically learned during the training process. For each attribute, its representation vector $v^{A}$ is extracted from all the embedding vectors of its input feature collection $\{v^{F}_{1},...,v^{F}_{N^A}\}$ with a self-attention mechanism \cite{lin2017structured}:
\begin{equation}
\begin{aligned} 
& e^{F}_{i} = (u^{F})^{\intercal}{\rm tanh}(\textbf{W}^{F}v^{F}_{i} + b^{F}), \\
& a^{F}_{i} = \frac{{\rm exp}(e^{F}_{i})}{\sum_{j=1}^{N^{A}}{\rm exp}(e^{F}_{j})},\\
& v^{A} = \sum_{i=1}^{N^{A}}a^{F}_{i}v_{i}^{F},
\end{aligned} 
\end{equation}
where $\textbf{W}^{F}$, $b^{F}$ and $u^{F}$ are trainable weight matrix, bias vector and attention vector, respectively.
$N^A$ denotes the number of the input features belonging to the attribute. $v_i^F$, $e_i^F$ and $a_i^F$ denote the embedding vector, attention score, and normalized attention score of the $i$th feature, respectively.

\subsubsection{Instance Representation}
Based on the attribute vectors, we construct the higher-level representation for each input instance from two channels, i.e., self representation and relative representation against the previous data block.

The self representation vector, $v^{S}$, is extracted from the instance's attribute representations $\{v^{A}_{1},...,v^{A}_{N^I}\}$: 
\begin{equation}
\begin{aligned} 
& e^{A}_{i} = (u^{A})^{\intercal}{\rm tanh}(\textbf{W}^{A}v^{A}_{i} + b^{A}), \\
& a^{A}_{i} = \frac{{\rm exp}(e^{A}_{i})}{\sum_{j=1}^{N^{I}}{\rm exp}(e^{A}_{j})},\\
& v^{S} = \sum_{i=1}^{N^{I}}a^{A}_{i}v_{i}^{A},
\end{aligned} 
\end{equation}
where $\textbf{W}^{A}$, $b^{A}$ and $u^{A}$ are trainable weight matrix, bias vector and attention vector, respectively.
$N^I$ denotes the number of attributes.
$v_i^A$, $e_i^A$ and $a_i^A$ denote the embedding vector, attention score, and normalized attention score of the $i$th attribute, respectively.

The relative representation vector, $v^{R}$, is calculated by comparing the instance's attributes with the previous block vector. 
It is designed to enable the model to extract instance's relative patterns against the larger-scale collective patterns of the data: 
\begin{equation}\label{eq:repvr}
\begin{aligned} 
& e^{R}_{i} = ({u^{R}})^{\intercal}{\rm tanh}(\textbf{W}^{R}[f(v^{A}_{i}),v^{Mem}]+b^{R}),\\
& a^{R}_{i} = \frac{{\rm exp}(e^{R}_{i})}{\sum_{j=1}^{N^{I}}{\rm exp}(e^{R}_{j})},\\
& v^{R} = \sum_{i=1}^{N^{I}}a^{R}_{i}v^{A}_{i},
\end{aligned} 
\end{equation}
where the square brackets $[\,,\,]$ denotes the concatenation operation. The transformation function $f(\cdot)$ is the Leaky ReLU activation function. $\textbf{W}^{R}$, $b^{R}$ and $u^{R}$ are trainable weight matrix, bias vector and attention vector, respectively.
$e_i^A$ and $a_i^A$ are the attention score and normalized attention score of the $i$th attribute, respectively. $v^{Mem}$ is the memory vector from the previous data block and its calculation will be described in the next section. %subsection \ref{subsubsec:sl}.

For the output of this module, we concatenate the two latent vectors to form the final representation vector of the instance:
\begin{equation}\label{eq:repvi}
v^{I} = {\rm batch\_norm}([v^{S},v^{R}]).
\end{equation}

\subsubsection{Block Representation} \label{subsubsec:sl}

Furthermore, we go beyond instance level and implement a Recurrent Neural Network (RNN) cell to capture the long-term collective patterns of the sequential instances. For each data block, the representation can be calculated as:

\begin{equation}\label{eq:repvb}
v^{B}_{i} = {\rm RNN}(f(v^{I}_{i}),v^{B}_{i-1}), i=1,...,N^{B},
\end{equation}
where $N^{B}$ is the instance number in the block. 

The last hidden state, denoted by $v^{B}$, contains the latest and most information about the data's collective patterns over time. For this reason, we use $v^{B}$ as a representation for the block to improve the block-level anomaly detection (will be revisited in the following paragraphs). Moreover, it is also used as both the memory vector $v^{Mem}$ (in Eq. \ref{eq:repvr}) and the initial hidden state for the next block.

\subsection{Adversarial Learning}
Following the idea of Adversarial Autoencoder, we build an adversarial learning architecture to learn the intrinsic patterns of the training data for both instance and block levels. On one hand, the encoder-decoder generator part learns to generate resembled representations of the inputs. In this way, cycle consistency \cite{zenati2018adversarially} is enforced in the latent space. On the other hand, the discriminator tries to distinguish the real and resembled representations. 

\subsubsection{Instance Generator}
We use an autoencoder to generate resembled instance vectors.
The autoencoder first encodes the instance representation vector $v^{I}$ into a hidden space, and subsequently decodes it back to reconstruct a representation vector $v^{I*}$:
\begin{equation}\label{eq:geni}
\begin{aligned} 
& h^{enc} = \textbf{W}^{enc}(f(v^{I})+\Delta)+b^{enc}, \\
& v^{I*} = f(\textbf{W}^{dec}f(h^{enc})+b^{dec})-\Delta,
\end{aligned} 
\end{equation}
where $\textbf{W}$s and $b$s are the trainable weights and biases, respectively.

The performance of autoencoder is vulnerable to the noise in training data \cite{zhou2017anomaly}. In order to get a more robust model, we add a standard Multivariate Gaussian random noise $\Delta \sim \mathcal{N}_{d}(0,E)$ into the encode-decode process, where $E$ is the identity matrix and $d$ is the dimension of instance vector.

\subsubsection{Block Generator}
To introduce adversarial learning for long-term patterns, we also reconstruct a resembled vector per block for $v^{B}$:
\begin{equation}\label{eq:genb}
v^{B*}_{i} = {\rm RNN}(f(v^{I*}_{i}),v^{B*}_{i-1}), i=1,...,N^{B}.
\end{equation}
To be consistent with the calculation of real block vector $v^{B}$, here,
we don't train the weight and bias for the RNN cell, but directly copy the values of the corresponding parameters used for Eq. \ref{eq:repvb}. Similarly, the last hidden state is taken as the final resembled vector of the current block, denoted by $v^{B*}$.

\subsubsection{Discriminator}
Following the standard setting of binary classification, we build two one-layer neural network classifiers for both instance and block levels: 
\begin{equation} \label{eq:yi}
\begin{aligned} 
& \hat{y}^{I}= \sigma(\textbf{W}^{I}x^{I}+b^{I}) \mbox{ with } x^{I} \in \{v^{I},v^{I*}\} 
\end{aligned} 
\end{equation}
and 
\begin{equation} \label{eq:yb}
\begin{aligned} 
& \hat{y}^{B}= \sigma(\textbf{W}^{B}x^{B}+b^{B}) \mbox{ with } x^{B} \in \{v^{B},v^{B*}\},
\end{aligned} 
\end{equation}
where $\textbf{W}$s and $b$s are the trainable weights and biases, respectively. $\sigma(\cdot)$ denote sigmoid activation function.

\subsection{Training and Inference}
\subsubsection{Training}
In the training stage, we assume all training data are normal data to train our model in the unsupervised manner. As generative adversarial training is hard to converge, we don't minimize the generator loss and discriminator loss at the same time. Instead, we minimize the two losses in an alternative process: first holding the discriminator loss $\mathcal{L}_{\mathcal{D}}$ and minimizing generator loss $\mathcal{L}_{\mathcal{G}}$ for several steps, and then minimizing discriminator $\mathcal{L}_{\mathcal{D}}$ with the generator loss $\mathcal{L}_{\mathcal{G}}$ being held. 

For the generator loss, we use the sigmoid cross entropy \cite{creswell2017denoising} between real and resembled vectors 
\begin{equation} %\label{eq:yi}
\mathcal{L}_{\mathcal{G}}^{I} = {\sigma(v^{I})}^{\intercal}log( \sigma(v^{I*})) + (\mathbbm{1}-\sigma(v^{I}))^{\intercal} log (\mathbbm{1}-\sigma(v^{I*}))
\end{equation}
as the instance generator loss, and
\begin{equation}%\label{eq:yb}
\mathcal{L}_{\mathcal{G}}^{B} = {\sigma(v^{B})}^{\intercal}log( \sigma(v^{B*})) + (\mathbbm{1}-\sigma(v^{B}))^{\intercal} log (\mathbbm{1}-\sigma(v^{B*}))
\end{equation}
as the block generator loss.
The total generator loss to minimize is the sum of block generator loss and the average of its $N^{B}$ instance generator losses:
\begin{equation} %\label{eq:yi}
\mathcal{L}_{\mathcal{G}}=
\frac{1}{N^{B}}\sum_{i=1}^{N^{B}}\mathcal{L}_{\mathcal{G},i}^{I}+\mathcal{L}_{\mathcal{G}}^{B}.
\end{equation}

For the discriminator loss, under standard setting of binary classification, we also use cross entropy based on the output of Equations \ref{eq:yi} and \ref{eq:yb}
\begin{equation}\label{eq:ascorei}
\mathcal{L}_{\mathcal{D}}^{I} = y^{I}log(\hat{y}^{I}) + (1-y^{I})log (1-\hat{y}^{I})
\end{equation}
as the instance discriminator loss, and
\begin{equation}\label{eq:ascoreb}
\mathcal{L}_{\mathcal{D}}^{B} = y^{B}log(\hat{y}^{B}) + (1-y^{B})log (1-\hat{y}^{B})
\end{equation}
as the block discriminator loss. $y \in \{y^I,y^B\}$ is defined as: $y=1$ for real vectors and $y=0$ for resemble vectors.
In each optimization step, the total discriminator loss of a data block to minimize is:
\begin{equation} 
\mathcal{L}_{\mathcal{D}}= \frac{1}{N^{B}}\sum_{i=1}^{N^{B}}\mathcal{L}_{\mathcal{D},i}^{I}+\mathcal{L}_{\mathcal{D}}^{B}.
\end{equation}

\subsubsection{Inference}
For inference, we use compound loss as the output anomaly score to measure the degree of abnormality. Because the model lowers down the total loss by learning to fit normal data patterns during training. Abnormal data will produce higher loss since the model fails to fit abnormal patterns.
The anomaly score of an instance is given by:
\begin{equation}\label{eq:asocrei}
\begin{aligned} 
z^{I} = \mathcal{L}_{\mathcal{G}}^{I} + \beta \cdot \mathcal{L}_{\mathcal{D}}^{I}.
\end{aligned} 
\end{equation}
The anomaly score for a block is calculated by including both the block-level losses and the average instance anomaly score within the block:
\begin{equation}\label{eq:asocreb}
\begin{aligned} 
z^{B} = \mathcal{L}_{\mathcal{G}}^{B} + \beta \cdot \mathcal{L}_{\mathcal{D}}^{B} + \gamma \cdot \frac{1}{N^{B}}\sum_{i=1}^{N^{B}} z^{I}_{i}.
\end{aligned} 
\end{equation}
Two weight parameters $\beta$ and $\gamma$ are introduced to balance the influences from the different terms.

%% file: experiment.tex
\section{Experiment Setup}

\subsection{Datasets}
Three datasets are utilized to illustrate the performance of the proposed method. Their statistics are shown in Table \ref{tab:dataset}, and more details are described in Appendix.

\textbf{Synthetic dataset}: 
The Synthetic data is generated by adding random noises to multi-dimensional zigzag signals of discrete integers. The anomalies are constructed by either randomly generating numbers or randomly copying training instances.

\textbf{Public dataset}\footnote{https://archive.ics.uci.edu/ml/datasets/Connect-4}:
It is a public dataset about positions in the `connect-4' game. We use the instances labelled with `win' as normal data and the instances labelled with `loss' as anomaly data. Note that there is no sequential relation within the data.

%\textbf{Industrial dataset}\footnote{The Industrial dataset will be published upon the acceptance of the paper.}:
\textbf{Industrial dataset}\footnote{The Industrial dataset is published at https://tianchi.aliyun.com/dataset/dataDetail?dataId=27665.}:
The Industrial dataset is constructed by user behavior data from a real-world online recommendation system, which is very important for many tasks, such as click-through rate prediction \cite{zhou2018deep,ma2018entire}. The instances are collected over 10 consecutive days, and stored in the order of timestamp. As illustrated in Figure \ref{fig:example}, each instance consists of multiple attributes about user's past behaviors, e.g., clicked items in the past 3 days, favorite brands in the past week, etc. Each attribute contains a group of categorical ids in representation of the corresponding items, brands, etc. It’s impractical to get a dataset with enough well labeled anomalies from real-world production. Thus, we mimic anomalies by simulating real conditions (attributes are polluted by errors in upstream data pipeline). The anomalies are generated by deleting the records of a random selected attribute, or replacing the records with random ids. 
\begin{table}[h]
% \scriptsize

\small
\centering

\begin{tabular}{ccccc} 
 \toprule
\textbf{Dataset}&\textbf{\#Dimension}& \textbf{\#Attribute}&\textbf{\#Normal}& \textbf{\#Anomaly}\\ \midrule
Synthetic& 30&3 & 10,000 & 1,000\\ 
Public& 192 &64 & 44,000 & 4,000 \\ 
Industrial &440,512 &8 & 783,000 &25,000\\
\bottomrule
\end{tabular}
% \vspace{-0.5em} 
\caption{Statistics of three datasets used in the paper. The dimension denotes the total number of all the distinct categorical feature IDs for the entire dataset.}
\label{tab:dataset}
% \vspace{-1em} 
\end{table} 
\begin{table}[h]
% \scriptsize

\small
\centering

\begin{tabular}{ccc}
 \toprule
\textbf{Method}&\textbf{Reference}& \textbf{Category}\\ \midrule
OCSVM\tablefootnote{https://scikit-learn.org/\label{fn:sklearn}}& \cite{Scholkopf:2001:ESH}&SVM\\
iForest\footref{fn:sklearn}& \cite{liu2012isolation}& Tree ensemble\\
RDA\tablefootnote{https://github.com/zc8340311/RobustAutoencoder}& \cite{zhou2017anomaly}&Autoencoder\\
OCNN\tablefootnote{https://github.com/raghavchalapathy/oc-nn}& \cite{chalapathy2018anomaly}&Deep SVM\\
ALAD\tablefootnote{https://github.com/houssamzenati/Efficient-GAN-Anomaly-Detection}& \cite{zenati2018adversarially} &GAN\\
GANomaly\tablefootnote{https://github.com/samet-akcay/ganomaly}& \cite{akcay2018ganomaly}&GAN\\
ALOCC\tablefootnote{https://github.com/khalooei/ALOCC-CVPR2018}& \cite{sabokrou2018adversarially} &GAN\\ 
\bottomrule
\end{tabular}
% \vspace{-1em} 
\caption{Baseline methods in this paper.}
\label{tab:baseline}
% \vspace{-1em} 
\end{table} 
 
\begin{table*}[h]
% \scriptsize
\small
\centering
\begin{tabular}{cccccccccc} 
 \toprule
 \multirow{2}{*}{ \textbf{Model}}&\multicolumn{3}{c}{\textbf{Synthetic Dataset}}&\multicolumn{3}{c}{\textbf{Public Dataset}}&\multicolumn{3}{c}{\textbf{Industrial Dataset}}\\
 \cmidrule(){2-10}
 & \textbf{Accuracy} & \textbf{F1-macro}& \textbf{AUROC} & \textbf{Accuracy} & \textbf{F1-macro}& \textbf{AUROC} & \textbf{Accuracy} & \textbf{F1-macro}& \textbf{AUROC} \\ \midrule 
OCSVM& 0.650 & 0.644 & 0.641 & 0.578 & 0.577 & 0.609 & 0.591 & 0.591 & 0.623 \\ 
iForest& 0.650 & 0.650 & 0.671 & 0.576 & 0.575 & 0.618 & 0.561 & 0.559& 0.589 \\ 
RDA& 0.655 & 0.652 & 0.704 & 0.564 & 0.563 & 0.564 & 0.530 & 0.530 & 0.538\\ 
OCNN& 0.550 & 0.549 & 0.533 & 0.578 & 0.577 & 0.577 & 0.624 & 0.584& 0.624 \\ 
ALAD& 0.664 & 0.664 & 0.705 & 0.650 & 0.650 & 0.706 & 0.618 & 0.618 & 0.661 \\ 
GANomaly& 0.646 & 0.644 & 0.703 & 0.676 & 0.676 & 0.709 & 0.610 & 0.610 & 0.657 \\ 
ALOCC& 0.638 & 0.637 & 0.692 & 0.682 & 0.681 & 0.703 & 0.612 & 0.611 & 0.648 \\ 
AMAD & \textbf{0.680}* & \textbf{	
0.681}* & \textbf{0.717}* & \textbf{0.700}* & \textbf{0.689}* & \textbf{0.730}* & \textbf{0.644}* & \textbf{0.643}* & \textbf{0.691}* \\
 \bottomrule
\end{tabular}
% \vspace{-1em} 
\caption{Results of instance anomaly detection on three datasets.
Symbol `*' highlights the cases where our model significantly beats the best baseline with $p$ value smaller than 0.01.}
\label{tab:instance}
% \vspace{-1em} 
\end{table*}

\begin{table*}[h]
% \scriptsize

\small
\centering
\begin{tabular}{cccccccccc} 
 \toprule
 \multirow{2}{*}{ \textbf{Model}}&\multicolumn{3}{c}{\textbf{Synthetic Dataset}}&\multicolumn{3}{c}{\textbf{Public Dataset}}&\multicolumn{3}{c}{\textbf{Industrial Dataset}}\\
 \cmidrule(){2-10}
 & \textbf{Accuracy} & \textbf{F1-macro}& \textbf{AUROC} & \textbf{Accuracy} & \textbf{F1-macro}& \textbf{AUROC} & \textbf{Accuracy} & \textbf{F1-macro}& \textbf{AUROC} \\ \midrule 
OCSVM& 0.543 & 0.528 & 0.555 & 0.580 & 0.573 & 0.562& 0.572 & 0.561 & 0.581 \\ 
iForest& 0.672 & 0.660 & 0.674 & 0.540 & 0.539 & 0.562 & 0.501 & 0.499 & 0.494 \\ 
RDA& 0.644 & 0.633 & 0.680 & 0.538 & 0.535& 0.543 & 0.520 & 0.508 & 0.499 \\ 
OCNN& 0.512 & 0.508 & 0.498 &0.572 & 0.554 & 0.561& 0.626 & 0.577 & 0.648 \\ 
ALAD& 0.600 & 0.599 & 0.629 & 0.655 & 0.650 & 0.683 & 0.600 & 0.596 & 0.620 \\ 
GANomaly& 0.563 & 0.555 & 0.574 & 0.652 &0.642 & 0.679 & 0.579 & 0.557 & 0.580 \\ 
ALOCC & 0.641& 0.625 & 0.640 & 0.512 & 0.512 & 0.523 & 0.581 & 0.576 & 0.605 \\ 
AMAD & \textbf{0.764}* & \textbf{0.767}* & \textbf{0.745}* & \textbf{0.660}* & \textbf{0.659}* & \textbf{0.746}* & \textbf{0.655}* & \textbf{0.655}* & \textbf{0.674}* \\
 \bottomrule
\end{tabular}
% \vspace{-1em} 
\caption{Results of block anomaly detection on three datasets. Symbol `*' highlights the cases where our model significantly beats the best baseline with $p$ value smaller than 0.01.}
\label{tab:block}
% \vspace{-1em} 
\end{table*}
For the Public dataset, the normal data are randomly split for either training or testing, while the testing normal data is mixed with the anomalous data to form the final testing set.
For the Synthetic and Industrial datasets, the former part (the majority) of the normal data is used for training, whereas the last small portion is used separately and mixed with anomaly samples for testing. To better cover the high-dimension space outside the normal data and test the models' performance more efficiently, we include a large ratio of anomalies for the test data. All the testing datasets have a half-to-half ratio of normal and anomalous instances. Unbalanced test set can be made by down sampling anomalies, and the corresponding performance metrics such as recall and precision can be calculated by adjusting the reported results with sample ratio. For block-level detection, we define that a testing data block is anomalous if more than 50\% instances in a block are anomalous. Otherwise the block is defined as normal.

\subsection{Settings}
\subsubsection{Parameters and Metrics}
We report the results with best hyper-parameters from grid search. We train our model using RMSProp optimizer with learning rate as 0.01. We set block size as 100, $\beta = 0.3$ and $\gamma = 0.05$ for calculating the compound anomaly score (Eqs. \ref{eq:asocrei} and \ref{eq:asocreb}).

In this paper, we report Accuracy, F1-score and AUROC as the metrics to evaluate model performance. With the anomaly scores of all testing data, we firstly calculate AUROC score by considering all possible thresholds and subsequently pick the optimal threshold as defined in \cite{habibzadeh2016determining}. Based on the optimal threshold, Accuracy and F1-score are calculated in the end.

\subsubsection{Baselines}
As listed in Table \ref{tab:baseline}, we selected 7 state-of-art methods for comparison with proposed model. These models only output anomaly score for each individual instance. We use the average score of the instances in a block as the block-level anomaly score. We also want to note that the baseline models are not originally designed for the complex high-dimensional categorical data such as our Industrial dataset. Therefore, for the Industrial dataset, we embed the instances into vectors ahead with Doc2Vec \cite{le2014distributed}, and use the embedded vectors of instances as the input for the baseline methods.

\section{Results}
In this section, we report the experimental results to demonstrate our approach's superiority over the other methods. Moreover, we present studies to verify the effects of the important characteristics of our model.

\subsection{Performance of the Full Model}
% \scriptsize
We run our model ten times and report the average evaluation results on instance-level anomaly detection and block-level anomaly detection in Tables \ref{tab:instance} and \ref{tab:block}. 
To verify our model's superiority, we calculate the performance differences between our model and the best baseline on each metric for all the runs, and apply a T-test to check whether the performance difference is significantly above 0 or not.

We can find that the GAN-based models perform best among the baselines, while our model outperforms all the baselines with respect to all the metrics and datasets.
Moreover, our model displays larger advantages for block-level detection. Clearly, the block-level anomaly detection gets more benefits from our unified multiscale approach.

\begin{table*}[h]
% \scriptsize

\small
\centering
\begin{tabular}{cccccccccc} 
 \toprule
 \multirow{2}{*}{ \textbf{Model}}&\multicolumn{3}{c}{\textbf{Synthetic Dataset}}&\multicolumn{3}{c}{\textbf{Public Dataset}}&\multicolumn{3}{c}{\textbf{Industrial Dataset}}\\
 \cmidrule(){2-10}
 & \textbf{Accuracy} & \textbf{F1-macro}& \textbf{AUROC} & \textbf{Accuracy} & \textbf{F1-macro}& \textbf{AUROC} & \textbf{Accuracy} & \textbf{F1-macro}& \textbf{AUROC} \\ \midrule 
-- Noise& -4.0 & -4.2 & -0.1 & -7.5 & -8.7 & -2.8 & -2.4 &-2.2 & -3.2 \\ 
-- RelRep& -5.9 & -6.3 & -5.9 &-1.7 &-1.3& -0.4& -4.4 & -6.9 & -8.5 \\ 
 \bottomrule
\end{tabular}
% \vspace{-1em} 
\caption{Instance-level performance differences between the ablated model and the full model. Results are scaled by a factor of 100.}
\label{tab:ablation-instance}
% \vspace{-1em} 
\end{table*}
\begin{table*}[h]
% \scriptsize

\small
\centering
\begin{tabular}{cccccccccc} 
 \toprule
 \multirow{2}{*}{ \textbf{Model}}&\multicolumn{3}{c}{\textbf{Synthetic Dataset}}&\multicolumn{3}{c}{\textbf{Public Dataset}}&\multicolumn{3}{c}{\textbf{Industrial Dataset}}\\
 \cmidrule(){2-10}
 & \textbf{Accuracy} & \textbf{F1-macro}& \textbf{AUROC} & \textbf{Accuracy} & \textbf{F1-macro}& \textbf{AUROC} & \textbf{Accuracy} & \textbf{F1-macro}& \textbf{AUROC} \\ \midrule 
-- Noise& -6.4 & -7.0 & -0.3 & -3.0 & -5.7 & -0.7 & -2.2 & -2.2 & -0.6 \\ 
-- RelRep& -14.2 & -14.8 & -2.6 & 0.0 & -1.1 & -0.6 & -3.9 & -7.0 & -1.9 \\ 
-- BlockLoss& -10.0 & -11.1 & -7.4 & -2.7& -2.1& -0.9&-6.2&-5.3& -4.4 \\ 
 \bottomrule
\end{tabular}
% \vspace{-1em} 
\caption{Block-level performance differences between the ablated model and the full model. Results are scaled by a factor of 100.}
\label{tab:ablation-block}
% \vspace{-1em} 
\end{table*}

\subsection{Random Noise in Autoencoder} 
%\zheng{don't need to mention RDA}
The performance of autoencoder can be deteriorated due to the noises in the training data. We add a random noise into the autoencoder (Eq. \ref{eq:geni}), to make it more robust.
To check the utility of the added noises, we retrain an ablated model by removing the $\Delta$ in Eq. \ref{eq:geni}. As shown by the results in the first rows (\textit{--Noise}) of Tables \ref{tab:ablation-instance} and \ref{tab:ablation-block}, adding noise clearly improves the generalization performance for all the tests.

\subsection{Relative Representation of Instance} \label{subsubsec:rr}
We introduce the relative representation $v^{R}$ (Eqs. \ref{eq:repvr} and \ref{eq:repvi}) to improve the pattern recognition. To justify this, we retrain an ablated model (\textit{--RelRep}) without this module, i.e., deleting Eq. \ref{eq:repvr} and removing $v^{R}$ from Eq. \ref{eq:repvi}. It leads a big drop in the performance, as shown in the second rows of Tables \ref{tab:ablation-instance} and \ref{tab:ablation-block}. 
Through comparing the instance with block representation, $v^{R}$ notably enriches the information extracted at high level.

\subsection{Block Loss for Block-Level Detection} \label{subsubsec:bl}
To testify the effect of the block loss for detecting anomalous blocks. In consistent with baseline approaches, we remove the first two terms in Eq. \ref{eq:asocreb}, and only use the average instance anomaly score.
As listed in the third row of Table \ref{tab:ablation-block} (\textit{--BlockLoss}), the performance drops drastically down to the level very close to the baselines! The block loss adds detection of the collective patterns to our model, which is of great importance for detecting block-level anomaly.

\subsection{Performance on Non-Sequential Data}

Different with the two time-evolving datasets (Synthetic and Industrial dataset), the Public dataset is used to test the models under a simpler scenario, i.e., categorical data without sequential patterns.
As expected, on this non-sequential data, the sequence-related components of our model has very weak impact on the performance (\textit{--RelRep} and \textit{--BlockLoss} in Tables \ref{tab:ablation-instance} and \ref{tab:ablation-block}). 
Even though, our model still outperforms all the baselines (Tables \ref{tab:ablation-instance} and \ref{tab:ablation-block}), which demonstrates the intrinsic superiority of the hierarchical representation learning structure of our model.

\begin{figure} 
% 	\advance\leftskip-1cm 
\hspace*{-0.5cm} 
	\centering
	\includegraphics[width=0.8\columnwidth]{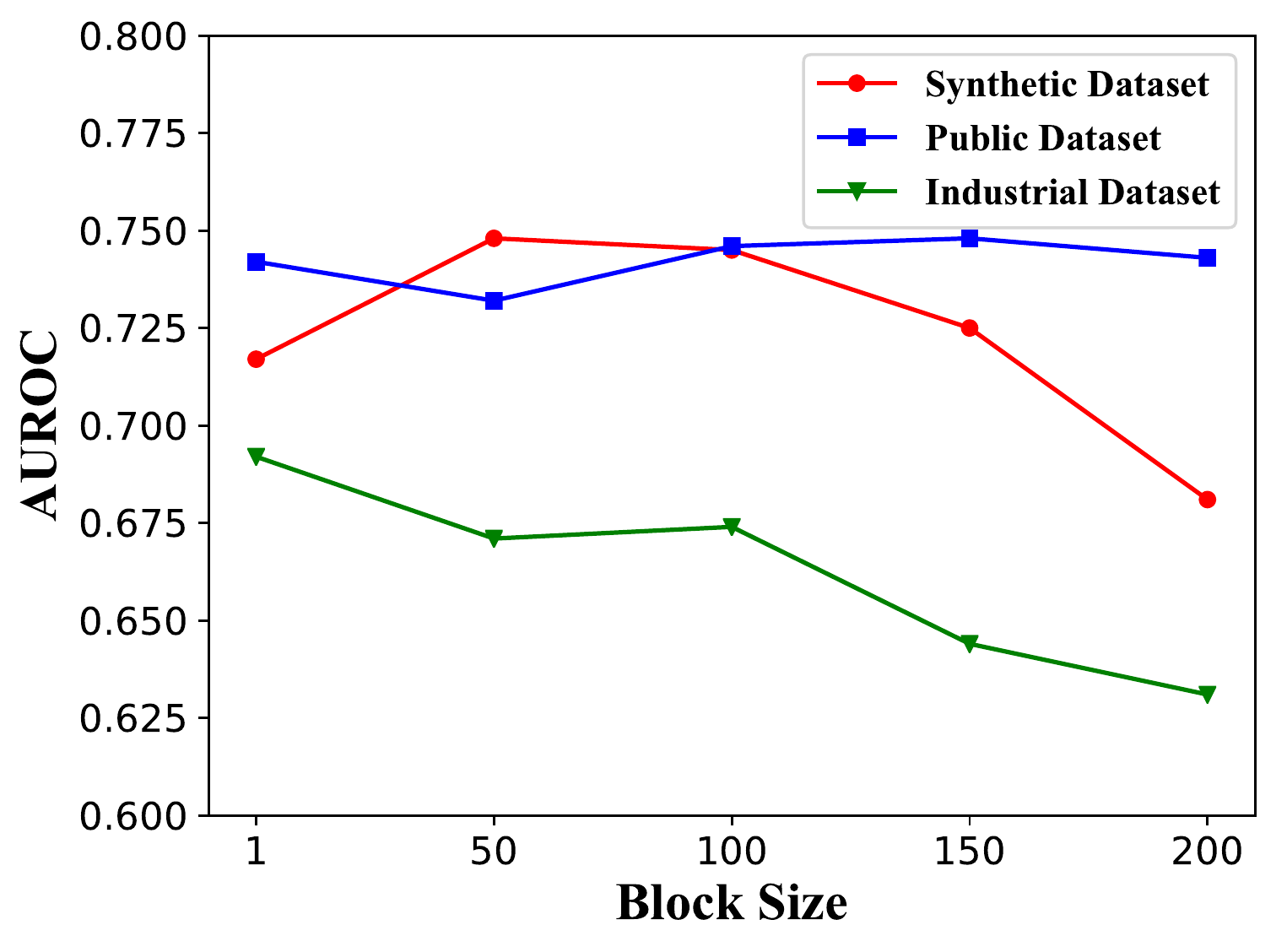}
% 	\vspace{-1em}
	\caption{The full model's performance of block level detection as a function of testing block size on the three datasets.}
\label{fig:block-size}
% \vspace{-1em} 
\end{figure}

\subsection{Block Size for Detection}
We also tested how the testing block size affects our model's detection performance (as shown in Figure \ref{fig:block-size}). 
For the Public dataset, there is no significant correlation between block size and the AUROC score, in consistent with the very weak effect of sequence-related components (as discussed in the last subsection).
For both Synthetic and Industrial datasets, we can see the performance drops for block size $>100$, where the testing block size exceeds the training block size. 
Also, for the left most case where block size $=1$, the AUROC scores are very close to the corresponding results in Table \ref{tab:instance}. Including block losses (Eq. \ref{eq:asocreb}) barely affect detection for individual instances.

%% file: conclusion.tex
\section{Summary and Discussion}
In this paper, we present AMAD, a multiscale Adversarial Autoencoder for anomaly detection at different levels on high-dimensional and time-evolving categorical data.
We demonstrate the effectiveness of our method by extensive experiments on datasets of different sizes and scenarios.

Real-world streaming data can have severe non-stationary data drift problem, which may require to keep the model updated in time through online incremental learning \cite{sahoo2017online}. In this work, we have trained our model in the setting of online incremental learning, i.e., processing the time-ordered data block by block. For future work, a more comprehensive study on online incremental learning will be followed.
Also, we will work on extending the approach to more complex industrial scenarios, i.e., multiscale detection cross large span of resolutions (from instance and block level to hour and day levels), large-scale distributed data process.

%We can adaptively capture the potential anomaly boundary drift via a time-aware attention. Extensive experiments on three different datasets show our model significantly outperforms the state-of-arts. However, as a common challenge for all GAN based models, the model performance are not always stable. In the future, we will explore more delicate mechanisms to tackle this problem such as considering statistic tests as a reinforcement guide.

%% file: appendix.tex
% \clearpage
% \newpage

\section{Appendix}

\subsection{Details of Datasets}

\textbf{Synthetic dataset}: 
We initialize the first instance with three categorical ids `0,10,20', and then generate the following instances by add 1 on each ID of the previous instance. If the ID surpasses 30, it will be subtracted by 30 and substituted by the remainder. 
Noises are introduced to the deterministic signals, by randomly selecting $10\%$ IDs and adding random noises $\in \{-1,1\}$ onto their original values. This process is repeated 220 times with a period of 50, generating 11000 normal instances.
We use the first 9000 normal instances as the training data and the remaining 2000 for testing. In the testing set, we randomly select 1000 instance and replace them with anomalies.
The anomalies are constructed by either randomly generating IDs or copying randomly selected training instances.

\textbf{Public dataset}: It is about positions in the `connect-4' game. Each feature in an instance refers to one of three choices (taken, not taken, blank) on a position. Each instance refers to a possible choice permutation. We use the instances labelled with `win' as normal data and the instances labelled with `loss' as anomaly data. We randomly select 40000 normal instance for training, while randomly mix the other 4000 with the 4000 anomalies to form the testing set.

\textbf{Industrial dataset}: 
The Industrial dataset is constructed by user behavior data from our online recommendation system. The user behavior record is updated according to user's latest behaviors.
Whenever the system receives an impression request from a user, a instance is generated.
The data is collected over 10 consecutive days, and stored in the order of timestamp.
All the real-world data is assumed to be normal. We use the first 758000 normal instances as the training data and the remaining 50000 for testing. In the testing set, we randomly select 25000 instances and replace them with anomalies.
The anomalies are generated by 
deleting the records under a random selected attribute, replacing the records with random IDs.